\documentclass[letterpaper, 10 pt, conference]{ieeeconf}

\IEEEoverridecommandlockouts
\usepackage{xspace}
\usepackage{amssymb,paralist,epsfig,standalone,bm,placeins}  
\usepackage{graphicx} 
\usepackage{xcolor}
\usepackage{multirow}
\usepackage{makecell}
\usepackage{multicol}
\usepackage[utf8]{inputenc}
\usepackage{amsmath,amsfonts,booktabs,cite} 
\usepackage{siunitx,textcomp} 
\usepackage[hidelinks]{hyperref} 
\let\labelindent\relax
\usepackage[inline]{enumitem} 
\usepackage[nolist]{acronym} 
\usepackage{caption}
\usepackage{tikz,standalone,pgfplots}
\usepackage{subfig}
\usetikzlibrary{patterns}
\usepackage{pgfplots, pgfplotstable}
\graphicspath{{figures_tex/}}

\usepackage{changes}

\pgfplotsset{compat=1.12}

\newacro{mss}[MSS]{Mass-Spring System}
\newacro{pbd}[PBD]{Position-based Dynamics}
\newacro{fem}[FEM]{Finite Element Method}
\newacro{dnn}[DNN]{Deep Neural Network}
\newacro{fcn}[FCN]{fully-convolutional network}

\newcommand{\figref}[1]{\hyperref[#1]{Fig.~\ref*{#1}}}
\newcommand{\tabref}[1]{\hyperref[#1]{Table~\ref*{#1}}}
\newcommand{\secref}[1]{\hyperref[#1]{Section~\ref*{#1}}}
\newcommand{\algoref}[1]{\hyperref[#1]{Algorithm~\ref*{#1}}}

\newcommand{\ra}[1]{\renewcommand{\arraystretch}{#1}}
\newcommand{\tbs}[1]{\renewcommand{\tabcolsep}{#1pt}}

\def\shake{\textit{shake task}\xspace}
\def\twist{\textit{twist task}\xspace}

\def\panda{Franka Emika Panda\xspace}

\def\ie{, \textit{i.e.}, }

\def\etal{\textit{et al.} }

\def\egad{EGAD!}

\def\figvspace{\vspace{-1.2em}}
\def\methodname{\textit{Def-GG-CNN}\xspace}

\definecolor{findOptimalPartition}{HTML}{D7191C}
\definecolor{storeClusterComponent}{HTML}{FDAE61}
\definecolor{dbscan}{HTML}{ABDDA4}
\definecolor{constructCluster}{HTML}{2B83BA}

\title{\LARGE \bf
Deformation-Aware Data-Driven Grasp Synthesis
}

\author{Tran~Nguyen~Le, Jens~Lundell, Fares~J.~Abu-Dakka, Ville~Kyrki%
\thanks{This work was supported by Academy of Finland Strategic Research Council
grant 314180 and CHIST-ERA project IPALM (326304). We gratefully acknowledge the support of
NVIDIA Corporation with the donation of the Titan Xp GPU used for this research.} \thanks{All authors are with
Intelligent Robotics Group at the Department of Electrical Engineering and
Automation, School of Electrical Engineering, Aalto University, Finland.
\texttt{\{firstname.lastname\}{@}aalto.fi}}
\thanks{Preliminary results have been presented in DO-Sim workshop celebrated at Robotics: Science and Systems conference \cite{tran2021}.}
}

\begin{document}
\maketitle
\thispagestyle{empty}
\pagestyle{empty}


\begin{abstract}
 Grasp synthesis for 3D deformable objects remains a little-explored topic, most works aiming to minimize deformations. However, deformations are not necessarily harmful---humans are, for example, able to exploit deformations to generate new potential grasps. How to achieve that on a robot is though an open question. This paper proposes an approach that uses object stiffness information in addition to depth images for synthesizing high-quality grasps. We achieve this by incorporating object stiffness as an additional input to a state-of-the-art deep grasp planning network. We also curate a new synthetic dataset of grasps on objects of varying stiffness using the Isaac Gym simulator for training the network. We experimentally validate and compare our proposed approach against the case where we do not incorporate object stiffness on a total of 2800 grasps in simulation and 420 grasps on a real \panda. The experimental results show significant improvement in grasp success rate using the proposed approach on a wide range of objects with varying shapes, sizes, and stiffness. Furthermore, we demonstrate that the approach can generate different grasping strategies for different stiffness values, such as pinching for soft objects and caging for hard objects. Together, the results clearly show the value of incorporating stiffness information when grasping objects of varying stiffness.
\end{abstract}

\section{Introduction}
\label{sec:introduction}

In the last decade, advancement in robotic grasping has enabled robots to automatically grasp a never-before-seen range of objects. However, most of the works on grasp synthesis still assume specific object properties such as uniform friction or rigidity. These assumptions do not hold for multi-material \cite{tran_friction} or deformable objects and can lead to unsuccessful grasping in real-world scenarios. 


\begin{figure}[!t]
	\centering
	\subfloat[Unsuccessful grasp executed on a rigid triangular-shaped object.]{
		\label{subfig:titlehard}
		\includegraphics[width=.9\linewidth]{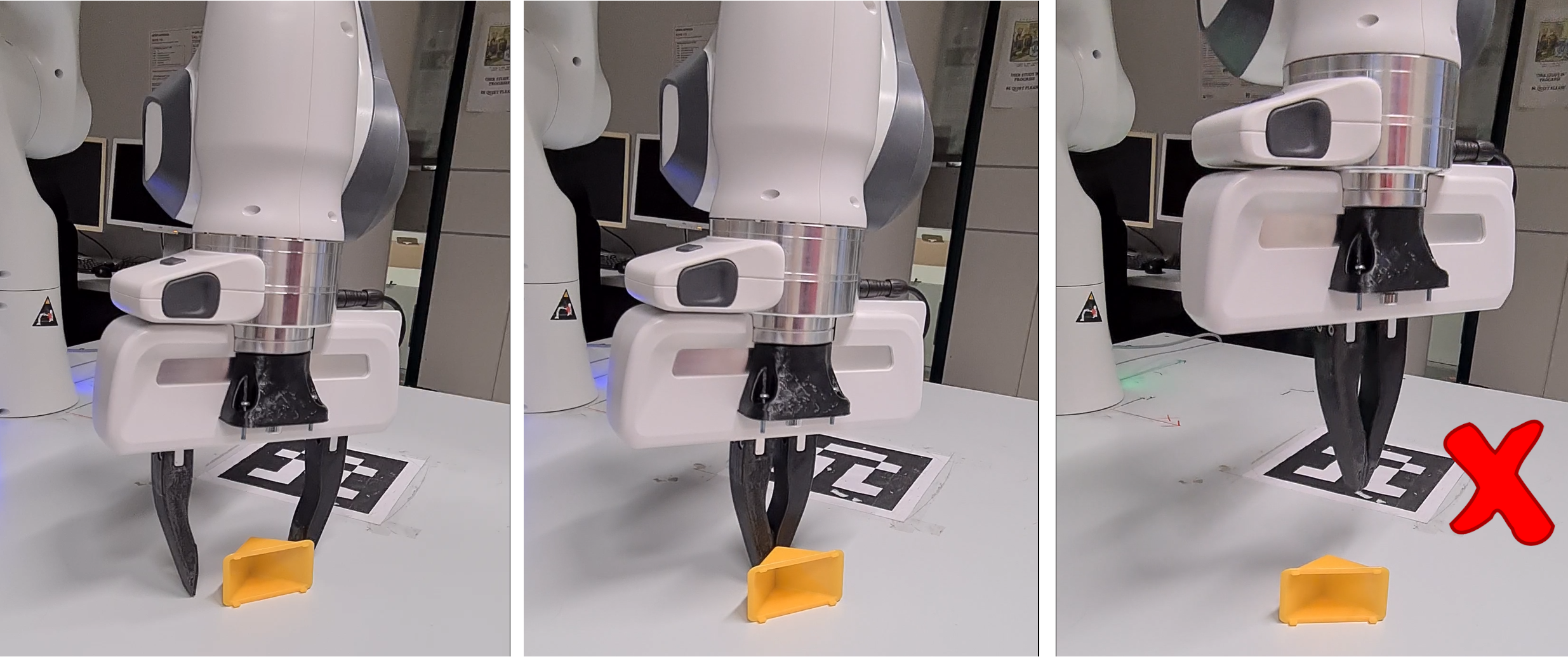} } \\
	\subfloat[Successful grasp executed on a deformable triangular-shaped object.]{
		\label{subfig:titlesoft}
		\includegraphics[width=.9\linewidth]{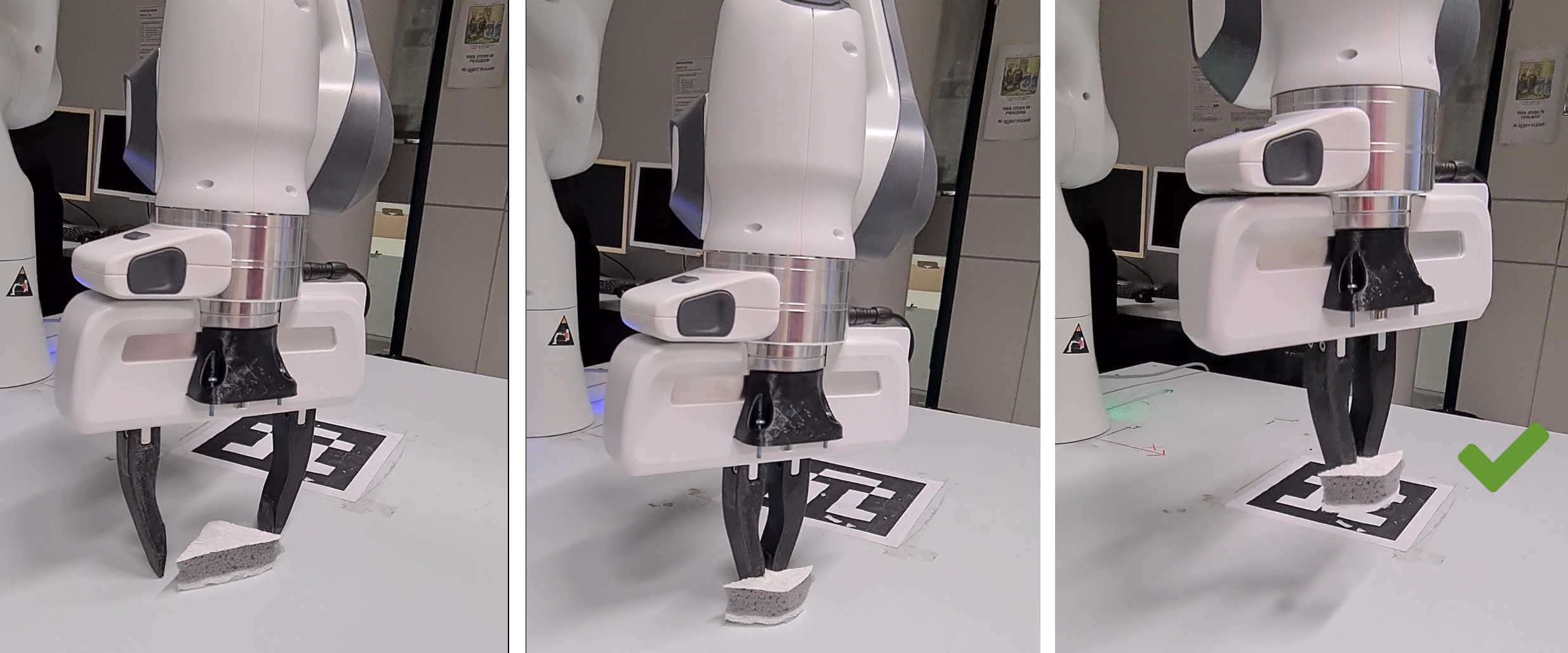} } 
	\caption{The robot executes the same grasp candidate on two objects with similar shapes but different stiffness. While the rigidity of the object tried to push itself out of the gripper (a), the deformation of the deformable object in (b) gently conformed to the shape of the gripper that leads to a successful grasp.}
	\label{fig:title}
	\vspace{-0.5cm}
\end{figure}

Grasping non-rigid objects, on the other hand, is difficult because objects deform under interaction forces meaning that the 3-D contact locations also depend on the forces exerted on the object. Furthermore, the effect of the deformation varies across deformable objects and tasks. In some scenarios, such as grasping a water bottle, it is useful to minimize the object's deformation not to dislodge the liquid. While for other objects, such as the triangular-shaped object shown in \figref{fig:title} one can take advantage of the deformation to grasp them successfully. To date, most of the existing works only focus on minimizing the object deformation \cite{alt_minimize,xu_minimalwork,pan_minimizedeform,delgado_minimize,soft_minimize}. Although there exist few works that take advantage of the deformation \cite{lin_v,lin_feel3d}, they mainly focus on proposing control strategies given an initial grasp configuration. Thus, it is still an open question of how object stiffness affects the choice of grasp configuration and how to harness object deformation to generate better grasps. 


To address the aforementioned open issues, we propose to incorporate stiffness as an additional input to a state-of-the-art deep grasp planning pipeline as shown in \figref{fig:pipeline}. Our system generates a grasp candidate and corresponding grasp quality for every pixel given an input depth image and a \textit{stiffness image}. The model outputs can be reprojected into 3D space when combined with depth information, allowing a robot to execute a generated grasp in the real world.


We qualitatively evaluated the proposed grasp synthesis method on a \panda robot in simulation and the real world by comparing it to a method that ignores stiffness. In the simulator, we evaluated over 2800 grasps on a shake and twist task, measuring, respectively, the grasp's robustness to linear and angular disturbances. In the real world, we measured the grasp success rate of 420 grasps on 14 objects under three different cases: with stiffness information, assuming all objects are rigid, and assuming all objects are deformable. In both simulation and the real world, our proposed approach demonstrates an improvement in grasp success rate. Furthermore, the approach can generate different grasping strategies for different stiffness values, such as pinching for soft objects and caging for hard objects, even though no pinch grasps were included in the training data.

In summary, the main contributions of this paper are:
\begin{itemize}
    \item The first generative stiffness-aware deep grasping approach that adapts the grasp location depending on the object's stiffness.
    \item The first stiffness-dependent image-based grasping dataset consisting of labeled top-down grasps on objects with varying stiffnesses.
    \item A thorough empirical evaluation of the proposed method presenting, both in simulation and on real hardware, improvements in terms of grasp ranking and grasp success rate over a method that ignores stiffness. 
\end{itemize}

\section{Related work}
\label{sec:related_work}
To put our work in context, we next review three complementary viewpoints, grasping of deformable objects, data-driven grasp synthesis, and simulation of deformable object interactions.
\subsection{Grasping Deformable Objects}
Most recent research on deformable object manipulation has mainly focused on manipulating cloth items \cite{cloth_emd, cloth_imi,cloth_drl} and ropes \cite{rope_zhu,rope_bohg,rope_te}. Grasping deformable objects remains a sparsely explored research area \cite{defgraspsim} with the majority of works focusing on formulating quality metrics to quantify the goodness of a grasp \cite{wakamatsu_bfc,alt_minimize,xu_minimalwork,ken_deform} or proposing control strategies \cite{delgado_minimize,soft_minimize,lin_v,lin_feel3d}. 


Most approaches for grasping deformable objects aim to minimize the deformation. For a particular grasp, the minimization can be performed on-line by employing a control strategy that regulates the force at each contact \cite{delgado_minimize,soft_minimize}. To plan grasps, minimum deformation can be achieved by placing fingers at locations with maximal local stiffness, determined e.g.\ using simulation \cite{alt_minimize}. The deformation can also be integrated as an additional component of a wrench-space grasp quality metric \cite{xu_minimalwork}.

In contrast to minimizing deformation, some works have proposed to utilize the object deformation, similar to this paper. Analytical grasp planning approaches following this line of study include bounded force closure  \cite{wakamatsu_bfc} which guarantees force closure under a bounded external force, and deform closure \cite{ken_deform} which generalizes form closure to deformable objects with frictionless contact. In the on-line case, finger displacements can be regulated in order to retain force closure, originally proposed for planar objects \cite{lin_v} and later extended to 3D \cite{lin_feel3d} by using \ac{fem} to continuously model changes in shape and contact geometry during object lifting.

Although \cite{lin_v,lin_feel3d} utilized object deformations for stable grasping, they focus on regulating either force or displacement of the fingers given an initial grasp configuration. This is in contrast to our work which focuses on the choice of the grasp configuration by taking advantage of the object deformation. 

\subsection{Data-driven Grasp Synthesis}
Rapid advances in deep learning research recently have changed the paradigm of robotic grasping from analytical methods to data-driven ones. The main reason for this paradigm shift is that data-driven methods have been proved to be able to generate grasps that typically achieve a high grasp success rate on a wide range of objects in just a matter of seconds, much faster compared to analytical methods \cite{redmon15,mahler2017dex,morrison2018closing,vishal_deep,song_graspwild,jens19}. For example, Mahler \etal \cite{mahler2017dex} used a dataset consisting of
millions of synthetic antipodal top-grasp to train a Grasp Quality Convolutional Neural Network (GQ-CNN) model that computes the probability of success of grasps from depth images. The GQ-CNN was further improved through the use of on-policy data and a \ac{fcn} structure called FC-GQ-CNN \cite{fcn-gq-cnn}. The \ac{fcn} structure has recently been found to perform well in grasp synthesis \cite{fcn-gq-cnn,morrison2018closing,zeng_fcn,edward_fcn}, having the ability to generate dense, pixel-wise predictions for an input image efficiently. 

Although the aforementioned works achieve impressive results on rigid objects, none of them explicitly investigate the usability on deformable objects especially 3D deformable objects. In this work, we tackle this problem by incorporating object deformation into a state-of-the-art deep data-driven grasping planning pipeline. 

\subsection{Soft-body Simulation}
Data-driven grasping requires training data, which in this work we generate in simulation. Simulating dynamics of deformable objects relies heavily on their geometric representations, for instance, particle representation is a good choice for simulating the dynamics of fluids. Yin \etal \cite{yin_survey} presents three primary deformable object modelling approaches, \ac{mss}, \ac{pbd}, and \ac{fem}, and their limitations. In this work, we decided to use \ac{fem} because it is often used to model 3D objects such as food \cite{heiden2021disect} and tissues \cite{fem_tissue} and,  compared to other modeling approaches, offers a more physically accurate representation of a deformable object in a continuous domain at the expense of computational cost.  

Some robotic simulators that support \ac{fem} are PyBullet\cite{pybullet}, SOFA\cite{sofa}, and NVIDIA's recent version of the Isaac Gym simulator \cite{isaacgym_ofi}, which supports soft body simulation through the NVIDIA Flex backend. Among the aforementioned simulators, Isaac Gym is chosen as it combines the advantages of the other two. Specifically, similar to SOFA, Isaac Gym includes a co-rotational linear model for precision in modeling and simulating the object deformation under interaction. Furthermore, similar to PyBullet, the Isaac simulator also provides the capability to integrate robot-related functions, making it easier to build robotic applications. 
Huang \etal  \cite{defgraspsim} also provides a grasping framework to automatically perform and evaluate grasp tests on an arbitrary target object. We use this framework in our work to generate training data and to test grasps.

\section{Problem Formulation}

This work addresses the problem of generating antipodal grasps on unknown objects with different stiffnesses lying on a supporting surface. The goal is to calculate a grasp for each pixel in the depth image while taking into account object stiffness. More formally, we train a model $\mathcal{M}$ that takes as input a depth image $\mathbf{I_d}$ and a stiffness image $\mathbf{I_s}$, and produces a grasp map \textbf{G} that incorporates grasp quality and grasp parameters (orientation, gripper width) for grasps centered at each pixel in the input.
\begin{equation*}
    \mathcal{M}: (\mathbf{I_d},\mathbf{I_s}) \mapsto \textbf{G}.
\end{equation*}

To achieve this goal, we propose to use the \ac{dnn} in \figref{fig:pipeline} to map from depth and stiffness images to grasps \textbf{G} in the image, which we can easily transform to the real world using known coordinate transforms. 
\section{Method}
\label{sec:method}

\subsection{Network}
Our solution is based on the GG-CNN because it is orders of magnitude smaller than other recent grasping networks, thus it is faster to train and evaluate the network, while achieving state-of-the-art results in grasping rigid objects.

\begin{figure}[tbp]
    \centering
    \def\svgwidth{\linewidth}
     {\fontsize{9}{8}
    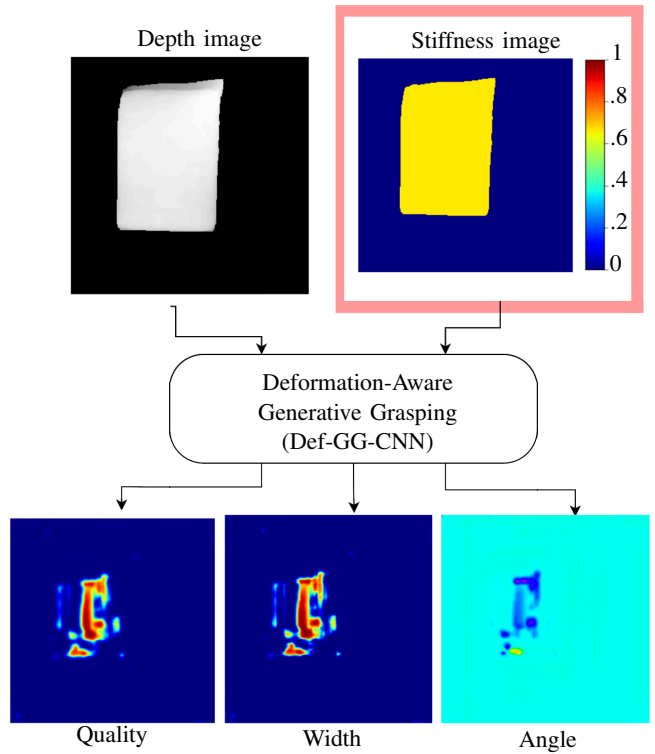}      
    \caption{The proposed pipeline where stiffness information (highlighted in red box) is incorporated.}
    \label{fig:pipeline}
    \vspace{-0.5em}
\end{figure}

We propose \methodname (\figref{fig:pipeline}), a fully convolutional network to synthesize grasps on objects with different stiffness including deformable ones. To enable \methodname to learn stiffness-dependent grasping strategies, it has, alongside the depth image, an additional stiffness image input. The stiffness image represents Young's modulus of the object at each pixel. The output of the network is the grasp map \textbf{G} that represents a grasp quality, and gripper parameters (orientation, gripper width) for each pixel of the depth image. The proposed network is trained with supervised learning on a synthetic dataset further explaned in \secref{sec:dataset}. 

\subsection{Grasp map representation}

Each pixel in the grasp map \textbf{G} represents a 4-dof grasp. We use the same representation of \textbf{G} as defined in \cite{morrison2018closing}. As shown in \figref{fig:pipeline}, the grasp map \textbf{G} consist of three images: grasp quality \textbf{Q}, orientation $\boldsymbol\phi$, and gripper width \textbf{W}.

\textbf{Q} denotes the quality of a grasp centered at each pixel. The quality of a grasp is a scalar value between [0,1], where the higher the value, the better the grasp. $\boldsymbol\phi$ is the orientation image, representing the pixel-wise orientation of a grasp around the image normal. Because an antipodal grasp is symmetric beyond 180 degrees, we limit the orientation between [$-\pi/2,~\pi/2$] radians. Finally, \textbf{W} is the width image that describes the pixel-wise gripper width from [0,~150] pixels. We transform the pixel-dependent gripper width to real-world units using the measured depth and the camera parameters.







\section{Dataset}
\label{sec:dataset}

To train \methodname, we need a dataset consisting of depth, stiffness, quality, orientation, and width images. To date, there exists no such dataset, and, thus, we opted to curate our own synthetic dataset. 

\begin{figure}[!t]
    \centering
    \def\svgwidth{\linewidth}
     {\fontsize{5}{8}
    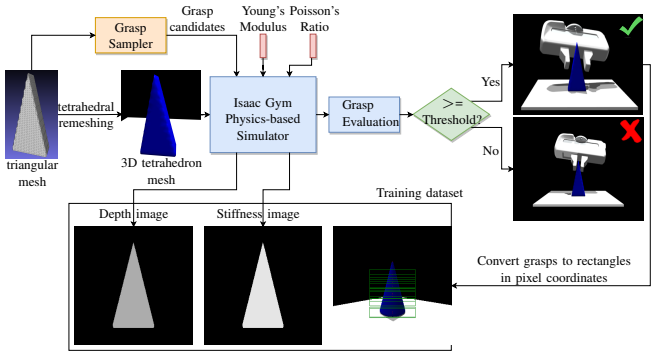} 
    \caption{The training data generation pipeline.}
    \label{fig:datagen}
    \vspace{-0.5cm}
\end{figure}

The pipeline of generating training data is visualized in \figref{fig:datagen}. We first convert the triangular mesh of an object into tetrahedral mesh using fTetWild \cite{ftetwild} and feed that tetrahedral mesh to the Isaac Gym simulator to enable its soft bodies simulation feature. The stiffness of an object can then be varied by adjusting the material parameters \ie Young's modulus and Poisson's Ratio. Then using the object triangular mesh, we sample grasp candidates which are later evaluated with Isaac Gym using proper quality metrics. Based on the performance of the grasps, we then label the grasps, convert them to the desired representation, and store them in the training dataset. 

\textbf{Depth and stiffness input:} We captured depth images of target objects with a virtual camera set to view the scene from top-down. To model variable object stiffness, four values of Young's modulus from $2\cdot 10^4$ to $2\cdot 10^9$ were used. The Young's modulus is normalized to [0,1] range and the corresponding stiffness value is assigned to every pixel in the stiffness image that the object occupies.

\textbf{Grasp candidates:} Grasps are sampled with an antipodal grasp sampler to obtain approximately 200 grasp candidates for each target object.  All grasps that collide with the mesh are filtered out, resulting in a final set of 25 to 40 collision-free grasps for each object.

\textbf{Quality metrics:} None of the standard grasp quality metrics, such as the Ferrari \& Canny $L_1$ metric \cite{ferraricanny}, are directly applicable for both rigid and deformable objects. As a quality metric, we use a \shake   which measures how easily an object is displaced in hand under linear accelerations. A higher metric means a better grasp as it indicates that a grasp can withstand higher accelerations. We use this metric to label a grasp as a positive or negative grasp by checking if the linear acceleration it can withstand is above or below a threshold. Specifically, after successfully lifting the object for each grasp candidate, we linearly increase the acceleration of the grasp along 16 directions until the gripper loses contact with the objects or reaches the upper acceleration limit, which is set to 50 $m/s^2$. Then we compute the average acceleration over all directions, and if this value is higher than the threshold of 25 $m/s^2$, we label the grasp candidate as a positive grasp.

\begin{figure}[!t]
    \centering
    \def\svgwidth{0.5\linewidth}
     {\color{white} \fontsize{8}{8}
\begingroup%
  \makeatletter%
  \providecommand\color[2][]{%
    \errmessage{(Inkscape) Color is used for the text in Inkscape, but the package 'color.sty' is not loaded}%
    \renewcommand\color[2][]{}%
  }%
  \providecommand\transparent[1]{%
    \errmessage{(Inkscape) Transparency is used (non-zero) for the text in Inkscape, but the package 'transparent.sty' is not loaded}%
    \renewcommand\transparent[1]{}%
  }%
  \providecommand\rotatebox[2]{#2}%
  \newcommand*\fsize{\dimexpr\f@size pt\relax}%
  \newcommand*\lineheight[1]{\fontsize{\fsize}{#1\fsize}\selectfont}%
  \ifx\svgwidth\undefined%
    \setlength{\unitlength}{459bp}%
    \ifx\svgscale\undefined%
      \relax%
    \else%
      \setlength{\unitlength}{\unitlength * \real{\svgscale}}%
    \fi%
  \else%
    \setlength{\unitlength}{\svgwidth}%
  \fi%
  \global\let\svgwidth\undefined%
  \global\let\svgscale\undefined%
  \makeatother%
  \begin{picture}(1,1.12418301)%
    \lineheight{1}%
    \setlength\tabcolsep{0pt}%
    \put(0,0){\includegraphics[width=\unitlength,page=1]{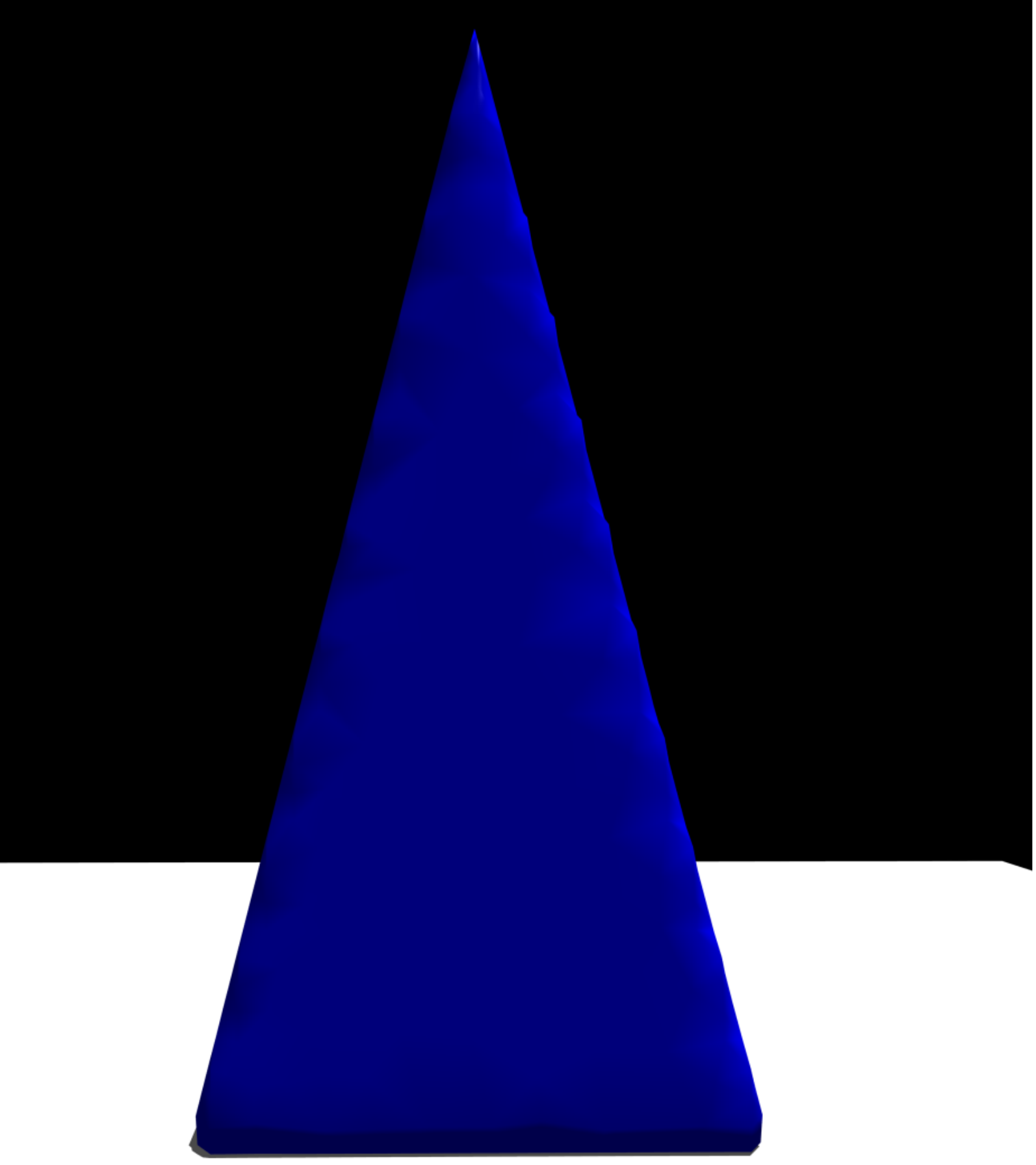}}%
    \put(0.7956094,0.91753466){\makebox(0,0)[t]{\lineheight{1.25}\smash{\begin{tabular}[t]{c}Grasp center\end{tabular}}}}%
    \put(0,0){\includegraphics[width=\unitlength,page=2]{grasprep.pdf}}%
    \put(0.80555007,1.02380852){\makebox(0,0)[t]{\lineheight{1.25}\smash{\begin{tabular}[t]{c}Grasp width\end{tabular}}}}%
    \put(0.79438273,0.81257537){\makebox(0,0)[t]{\lineheight{1.25}\smash{\begin{tabular}[t]{c}Grasp angle\end{tabular}}}}%
    \put(0.88367926,0.6389396){\makebox(0,0)[t]{\lineheight{1.25}\smash{\begin{tabular}[t]{c}Finger \\height\end{tabular}}}}%
  \end{picture}%
\endgroup%
}   
    \caption{A grasp is represented as a rectangle in 2D image plane.}
    \label{fig:grasprep}
    \vspace{-0.5cm}
\end{figure}

\textbf{Ground-truth grasp map:} To further simplify the data generation, we only use positively labeled grasps as ground-truth grasps to train the network. To generate the ground-truth grasps, we first transform all grasps to the image space. To do so, we represent the grasps as rectangles in the image as displayed in \figref{fig:grasprep}. Four parameters define the rectangles: grasp center, grasp orientation, grasp width, and finger height. Finally, we use the rectangles as image masks to generate ground-truth grasp maps \textbf{G}. Specifically, all pixels of the quality images \textbf{Q}, angle images $\boldsymbol\phi$, and width images \textbf{W} within the rectangle are set to the values given from the \shake. In contrast, all pixels outside the rectangle are set to invalid.  

\textbf{Training dataset:} As a training dataset, we generate and label grasps on 30 objects. The objects include 13 primitive objects provided in Isaac Gym, 5 objects from the YCB dataset \cite{ycb}, and 12 objects with adversarial geometry from the EGAD! dataset\cite{egad}. Because we set the stiffness for each object to four different values, the training set contains, in total, 120 objects. We use the \panda gripper model to execute grasps on objects in the simulator. To counteract the small size of the training set, we further augment the dataset with random crops, zooms, and rotations to create a set of 5400 depth and stiffness images with 27000 corresponding labeled grasp maps.  

\section{{Experiments and Results}}
\label{sec:exp_and_res}
\begin{table*}
    \centering
    \ra{1.3}\tbs{7}
    \caption{\label{tb:sim_exp_summary}Average grasp success rate (\%) on different stiffnesses for two different tasks. The higher the better.}
    \begin{tabular}{@{}l|cccc|cccc@{}}
        \toprule
        & \multicolumn{4}{c|}{Shake task} & \multicolumn{4}{c}{Twist task} \\
        \multicolumn{1}{c}{Stiffness ($\mathbf{E}$)} & \multicolumn{2}{c}{Common} & \multicolumn{2}{c|}{\egad{}} & \multicolumn{2}{c}{Common} & \multicolumn{2}{c}{\egad{}}  \\
        \cmidrule(lr){2-3} \cmidrule(lr){4-5} \cmidrule(lr){6-7} \cmidrule(lr){8-9} & With stiffness & No stiffness & With stiffness & No stiffness  & With stiffness & No stiffness & With stiffness & No stiffness\\
        \midrule
        $2\cdot 10^4$             & 75.4 & 22.8 &  67  & 40.8 & 65.7 & 20 & 69.5 & 51.3\\
        $2\cdot 10^5$             & 88.5 & 38.5 & 77.4 & 68.6 & 74.2 & 40 & 80 & 65.2\\
        $2\cdot 10^6$             & 88.5 & 48.5 & 77.4 & 69.6 & 71.4 & 42.8 & 75 & 64.4\\
        $2\cdot 10^9$             & 51.4 & 46.2 & 51.3 & 51.3 & 40   & 31.4 & 41.7 &41.7\\
        All stiffnesses                        & \textbf{75.9} & 40.3 &\textbf{68}& 57.6 &\textbf{63.5} & 34 & \textbf{67.3} & 55.6 \\
        \bottomrule
    \end{tabular}
    \label{tab:simresult}
 \figvspace{}
\end{table*}
The experiments address the following three questions:
\begin{itemize}
    \item Can \methodname synthesize high-quality grasps for deformable objects and would they succeed in simulation?
    \item Is \methodname robust against errors in the stiffness input?
    \item Can \methodname, trained purely on synthetic data, generate successful grasps in the real world?
    \item How does the stiffness information affect the choice of the grasp configuration?
\end{itemize}

\subsection{Grasping in Simulation}

\begin{figure}
    \centering
    \includegraphics[width=\linewidth]{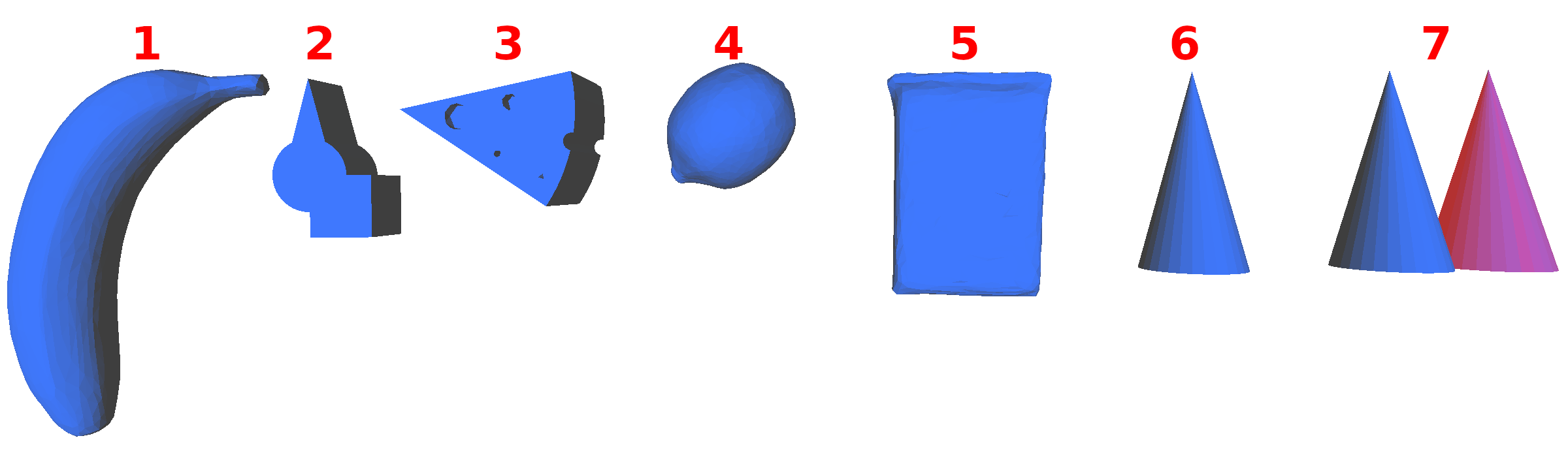}
    \caption{The seven common objects used in the
experiment. All objects are single-material except for object 7, where the stiffness of its red part can vary.}
    \label{fig:testobj}
    \vspace{-0.5cm}
\end{figure}

To investigate the quality of synthesized grasps in simulation, we evaluated the approach on two sets of objects: 7 common objects shown in \figref{fig:testobj}, and 28 adversarial objects from the recent EGAD! test dataset \cite{egad}. For each object and stiffness, we evaluated the five grasps with the highest quality on a \shake and \twist. While the \shake measures how easily an object slips out of the gripper under linear accelerations, the \twist measures that under angular accelerations. A grasp is successful if the object is in the gripper during the whole procedure, the grasp can withstand the linear acceleration limit of 25 $m/s^2$, and the angular acceleration limit of 500 $rad/s^2$. By doing this, we can quantify how the generated grasps behave under different disturbances. To demonstrate the importance and the effect of stiffness input, we compared the stiffness-aware grasps against grasps generated with an identical approach without stiffness information. In total, this amounts to 1400 grasps per method. 

\tabref{tab:simresult} shows the simulation result of both test sets. We can see that the proposed approach that takes stiffness input into account achieves a higher grasp success rate across all object sets and disturbances. It is noteworthy , however, that although we trained the network on the quality metrics from the \shake, the generated grasps also performed well on the \twist.

Focusing on the \shake results, the average grasp success rate using our approach over all stiffnesses compared to the baseline is 35\% higher on the Common objects and 11\% higher on EGAD! objects. Moreover, the performance of the baseline approach deteriorates significantly when moving from a high to a low value of Young's modulus. For instance, on the Common test set, the relative performance drop for the baseline approach when changing the Young's modulus from $2\cdot 10^6$ to $2\cdot 10^5$ is 10\%, and from $2\cdot 10^6$ to $2\cdot 10^4$ the drop is 26\%. This decline is much higher compared to the 0\% and 13\% drop using our approach. Similar performance differences are also observed for the EGAD! test set. The primary reason the baseline approach witnesses a higher performance drop is because it generates the same grasps for a target object regardless of its stiffness. Although the generated grasps often picked the objects successfully, they usually slip away from the gripper during the shake or twist task. In contrast, the network that took the stiffness input into account learned to avoid areas with a high probability of slippage, resulting in a higher grasp success rate. In addition, it is noteworthy that there is also some deterioration with the highest stiffness ($2\cdot 10^9$) for both approaches. The primary reason is that some objects in both test set have very complex shape which are extremely hard to grasp when they are rigid. This observation strengthen the idea of taking advantage of object deformation to successfully grasp complex-shaped objects.

We also evaluated the models on the multi-material object 7 shown in \figref{fig:testobj}, where the stiffness of the red part could differ from the blue part. The result showed that if we assumed object 7 is entirely rigid, the method could not generate any good grasps on it. However, if we assumed the red part was softer than the blue part, we could generate successful grasps that usually aimed for the softer area of the object. This simple example demonstrates the benefit of planning stiffness-aware grasps on irregular-shaped multi-material objects.

\subsection{Sensitivity Analysis}
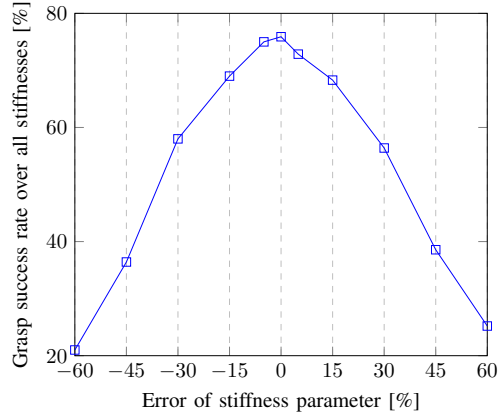
\begin{figure}[t]
	\centering
	\scalebox{0.8}{
\begin{tikzpicture}
\begin{axis}[
    xlabel={Error of stiffness parameter [\%]},
    ylabel={Grasp success rate over all stiffnesses [\%]},
    xmin=-60, xmax=60,
    ymin=20, ymax=80,
    xtick={-60,-45,-30,-15,0,15,30,45,60},
    ytick={20,40,60,80,100},
    legend pos=north west,
    xmajorgrids=true,
    grid style=dashed,
]

\addplot[
    color=blue,
    mark=square,
    ]
    coordinates {
    (-60,21)(-45,36.42)(-30,58)(-15,69)(-5,75)(0,75.9)(5,72.85)(15,68.3)(30,56.4)
    (45,38.57)(60,25.2)};
    
\end{axis}
\end{tikzpicture}
}
	\caption{Grasp success versus error in stiffness information. Grasp success deteriorates smoothly when the stiffness information is imprecise.}
	\label{fig:sensitivity_exp}
	\vspace{-1em}
\end{figure}
To examine the robustness of the results of our approach in the presence of uncertainty, we conducted a sensitivity analysis where we introduce uncertainty to the input stiffness images by varying the stiffness parameter \ie Young's modulus across an error range of [-60\%, +60\%]. We then evaluated the generated grasps on the Common test set under the \shake.

\figref{fig:sensitivity_exp} shows the result of the sensitivity analysis. We can see that the grasp success rate decreases consistently with increasing error, but that small errors in the range of 5-15\% have only limited negative effect on the performance. This experiment indicates that our method is robust against some errors in expected stiffness, which suggests potential for real-world application even when the stiffness is not precisely known. 

\subsection{Grasp Transfer to Physical Robot}
\begin{figure}
    \centering
    \includegraphics[width=.9\linewidth]{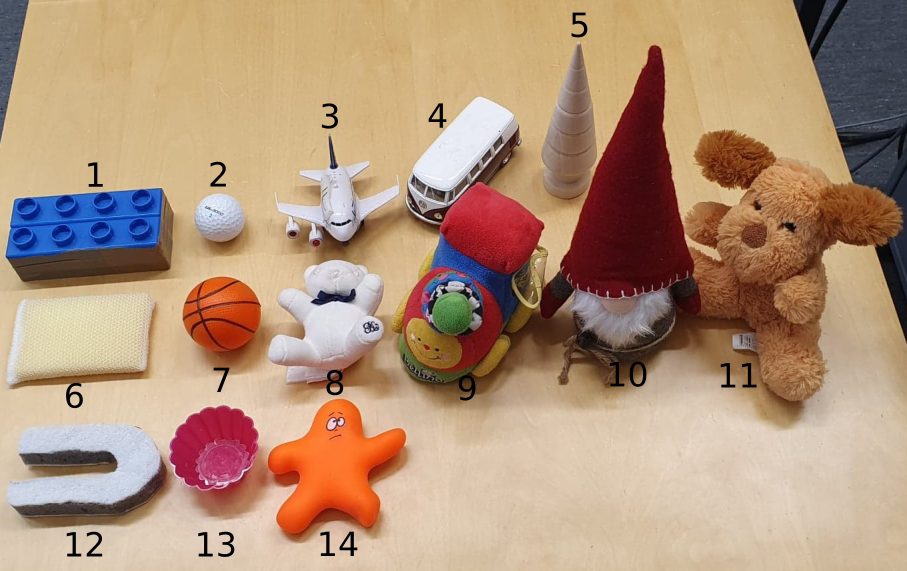}
    \caption{The 14 individually numbered objects used in the real experiment. The top, middle, and bottom rows include objects with high, medium, and low stiffness, respectively. }
    \label{fig:testobjreal}
    \vspace{-0.5cm}
\end{figure}
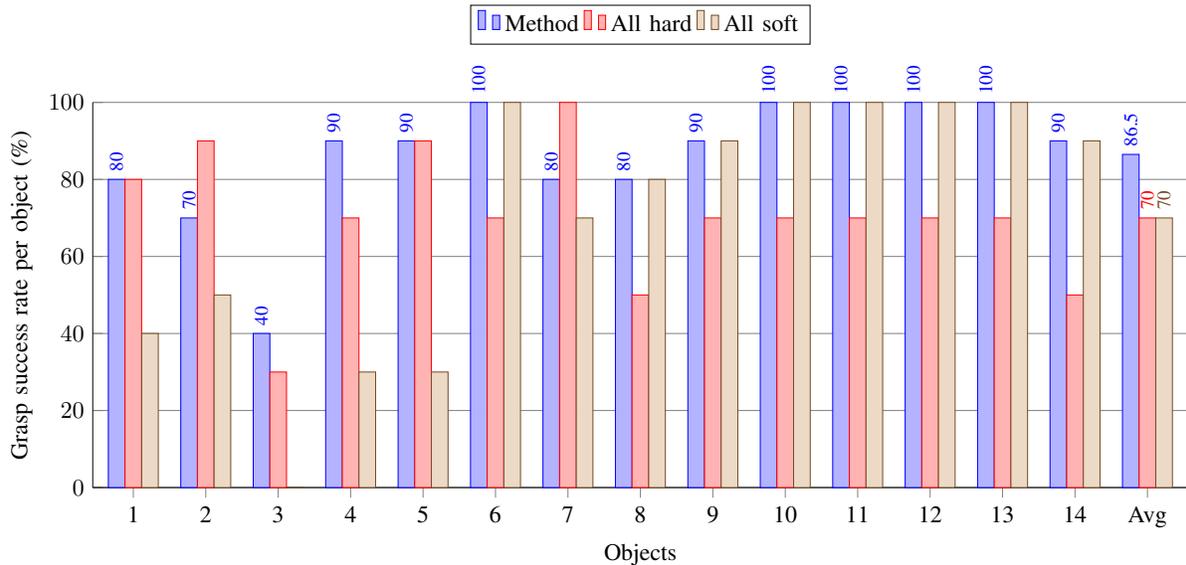
\begin{figure*}
	\centering
	\scalebox{0.9}{
\begin{tikzpicture}
\begin{axis}[
	ymin=0, ymax=100,
	ybar=0pt,
	bar width=7pt,
	x=1.07cm,
	xmin=1, xmax=Avg,
	enlarge x limits={.04},
	ylabel={Grasp success rate per object (\%)},
	xlabel={Objects},
	axis x line*=bottom,
	axis y line*=left,
	ymajorgrids,
	tick pos=left,
	y grid style={white!50.0!black},
	legend style={
		at={(0.5,1.15)},
		anchor=south,
		legend columns=-1,
	},
	symbolic x coords={1,2,3,4,5,6,7,8,9,10,11,12,13,14,Avg},
	xtick=data,
	nodes near coords,
	nodes near coords style={rotate=90, anchor=west,font=\footnotesize}, 
	point meta=explicit symbolic
]

\addplot+ coordinates{
	(1,80) [80]
	(2,70) [70]
	(3,40) [40]
	(4,90) [90]
	(5,90) [90]
	(6,100) [100]
	(7,80)  [80]
	(8,80) [80]
	(9,90) [90]
	(10,100) [100]
	(11,100) [100]
	(12,100) [100]
	(13,100) [100]	
	(14,90) [90]
	(Avg, 86.5)[86.5]
};

\addplot+ coordinates{
	(1,80) 
	(2,90) 
	(3,30) 
	(4,70)
	(5,90) 
	(6,70) 
	(7,100) 
	(8,50) 
	(9,70) 
	(10,70) 
	(11,70) 
	(12,70) 
	(13,70) 
	(14,50)
    (Avg, 70)[70]
};

\addplot+ coordinates{
	(1,40) 
	(2,50) 
	(3,0) 
	(4,30) 
	(5,30) 
	(6,100) 
	(7,70) 
	(8,80) 
	(9,90) 
	(10,100) 
	(11,100) 
	(12,100)
	(13,100)
	(14,90)
    (Avg, 70)[70]
};

\legend{Method, All hard, All soft}
\end{axis}
\end{tikzpicture}
}
	\vspace{-1em}
	\caption{Grasp success rate per object in three cases. The last column shows the average success rate for all three cases.}
	\label{fig:realresult}
	\vspace{-1em}
\end{figure*}

\begin{figure}[!t]
\centering
\subfloat[Pinch grasp on soft sponge $\mathbf{E} = 2\cdot 10^5$.]{
	\label{subfig:pinch}
	\includegraphics[width=.9\linewidth]{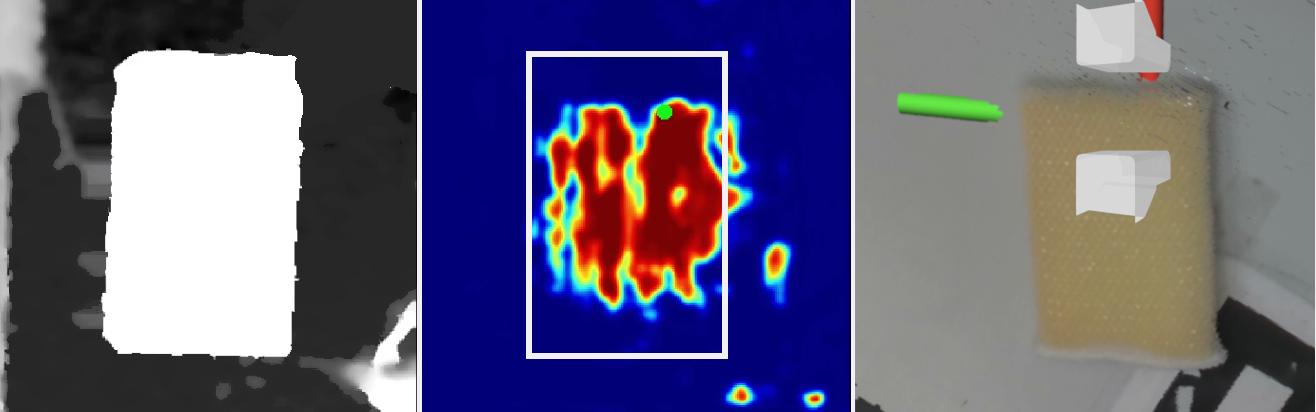} } \\
\subfloat[Cage grasp on rigid sponge $\mathbf{E} = 2\cdot 10^9$.]{
	\label{subfig:cage}
	\includegraphics[width=.9\linewidth]{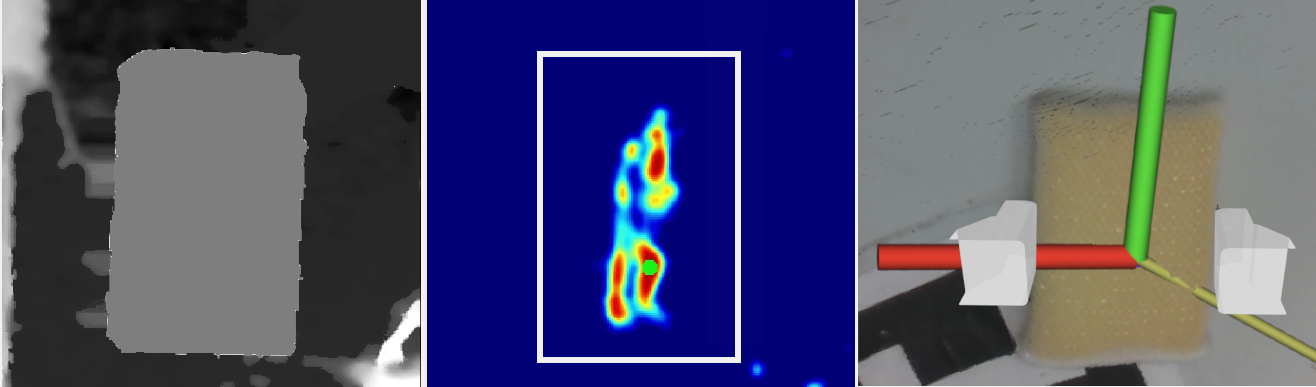} } 
\caption{Stiffness input image, along with grasp quality map and best synthesized grasp candidate indicated by two white fingers. For stiffness input image, the darker the color the stiffer the object is. For the grasp quality map, red indicates higher quality, and the green point denotes the best grasp.}
\label{fig:graspstrategies}
\vspace{-2em}
\end{figure}


To investigate how well the synthesized grasps perform in real world, we evaluated the grasp success rate on a \panda equipped with a parallel-jaw gripper. This allows us to study if grasps generated with the approach trained only on synthetic data transfer to real objects. 14 objects to grasp (\figref{fig:testobjreal}) were chosen as they represent a high variation in size, shape, and stiffness. 


We used an Intel RealSense D345 camera mounted to the robot's wrist to capture the RGB-D image. In addition to depth image, we also need to provide a stiffness image of the object. To do this, we segmented the object from the scene by subtracting the background and the table from the image and then assigning the same stiffness value to each pixel that the object occupied. We manually set the magnitude of stiffness for each object according to its perceived stiffness. The best grasp pose is then computed using the proposed method. 

The robot executed the best grasp by moving to a pre-grasp position approximately 25 mm above the grasp. Then, the robot moves linearly downwards until reaching the grasp pose or contact with the table is detected. From there, the robot closes its gripper, lifts the object, performs a predefined trajectory, and finally places an object at the goal position. A grasp is successful if the robot can pick the object and move it without dropping it. Otherwise, it is unsuccessful. 


To single out the effect stiffness input has on grasp performance for each object, we ran the experiments with three different stiffnesses: the correct one,  only high, only low. For each object and stiffness, we randomly placed it ten times and evaluated the best grasp candidate. In total, this setup amounts to 420 grasps on 14 objects. 


The experimental results are presented in \figref{fig:realresult}. The results show that with the correct stiffness information, the grasp success rate of the proposed approach is approximately 17\% higher than when the stiffness is assumed to be either high or low. This result indicates that grasps generated on rigid objects do not necessarily transfer successfully to deformable objects and vice versa. 

\begin{figure}[!t]
\centering
	\includegraphics[width=.9\linewidth]{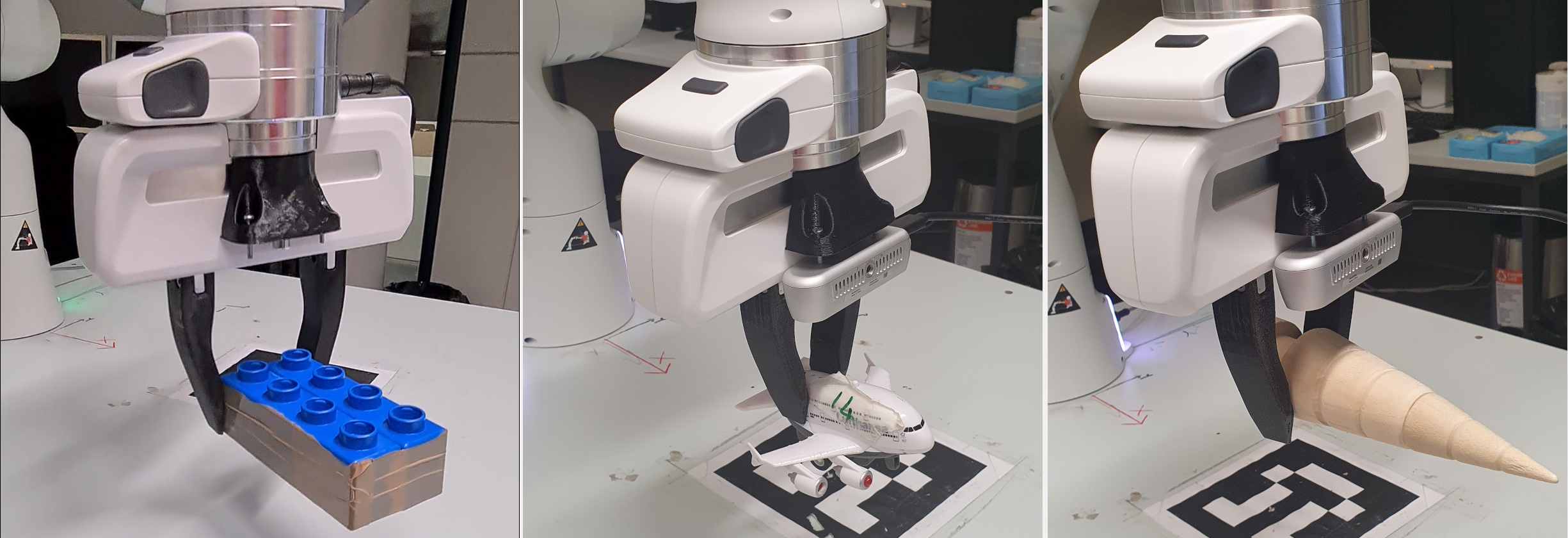} \\	
	\vspace{-0.5 em}
     \subfloat[]{
    	\includegraphics[width=.9\linewidth]{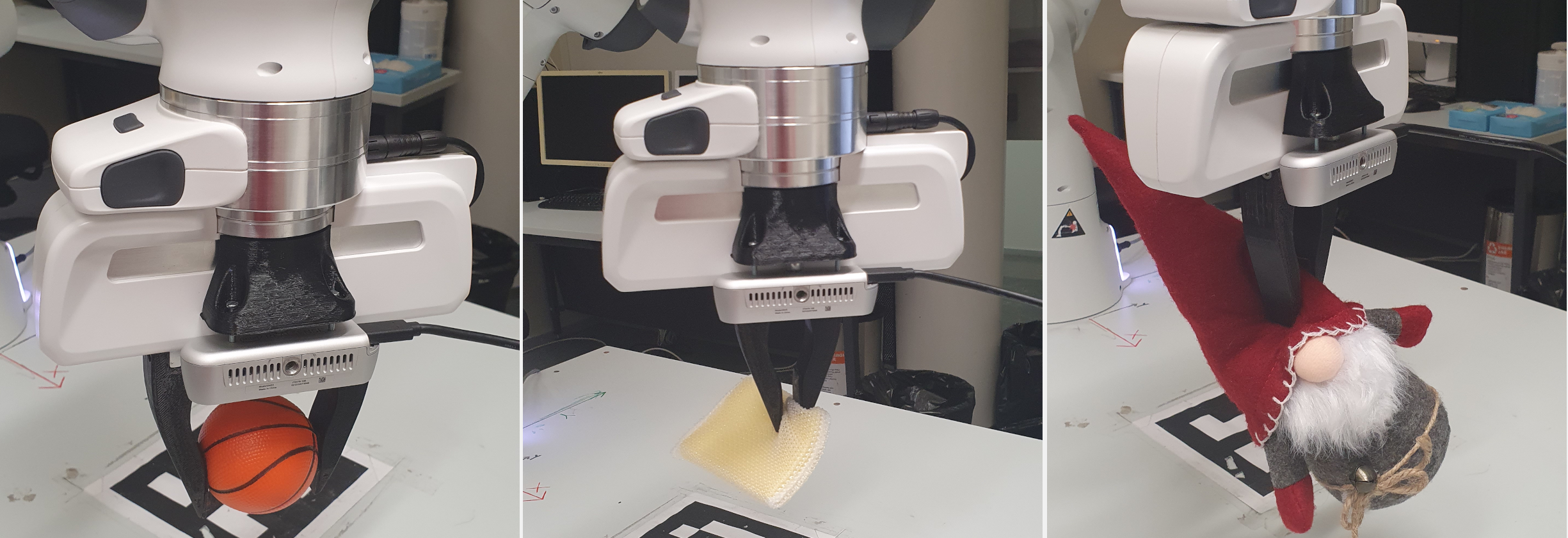} } \\
    	
	\includegraphics[width=.9\linewidth]{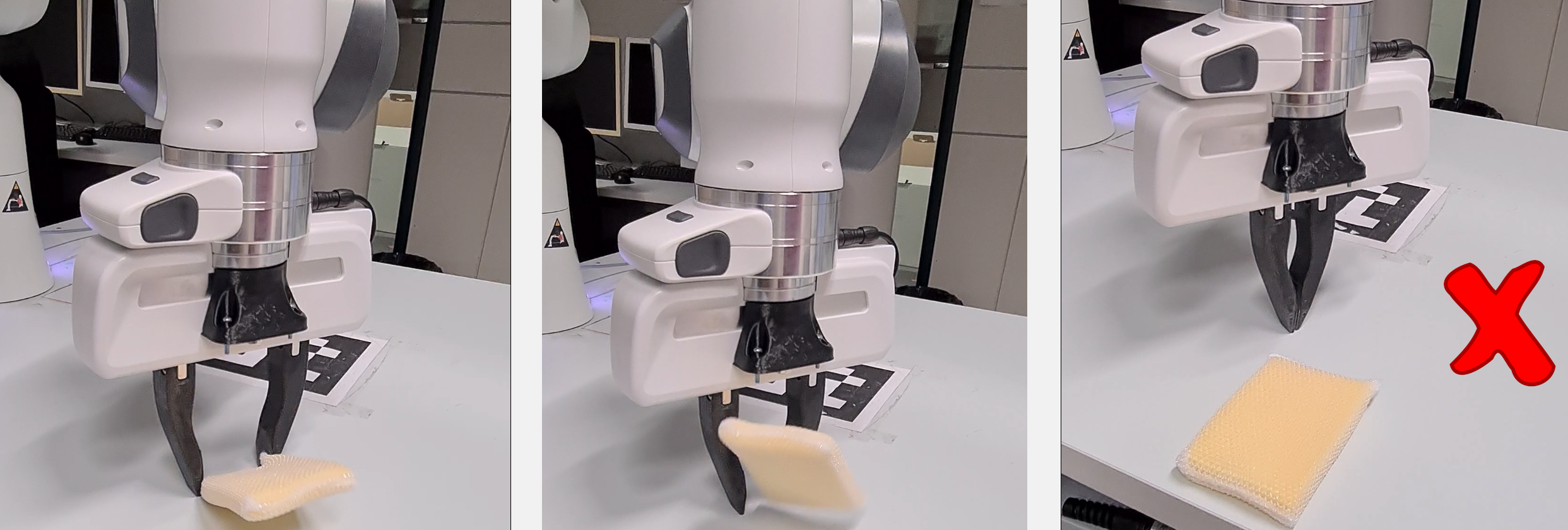}  \\
	\vspace{0.5 em}
	\includegraphics[width=.9\linewidth]{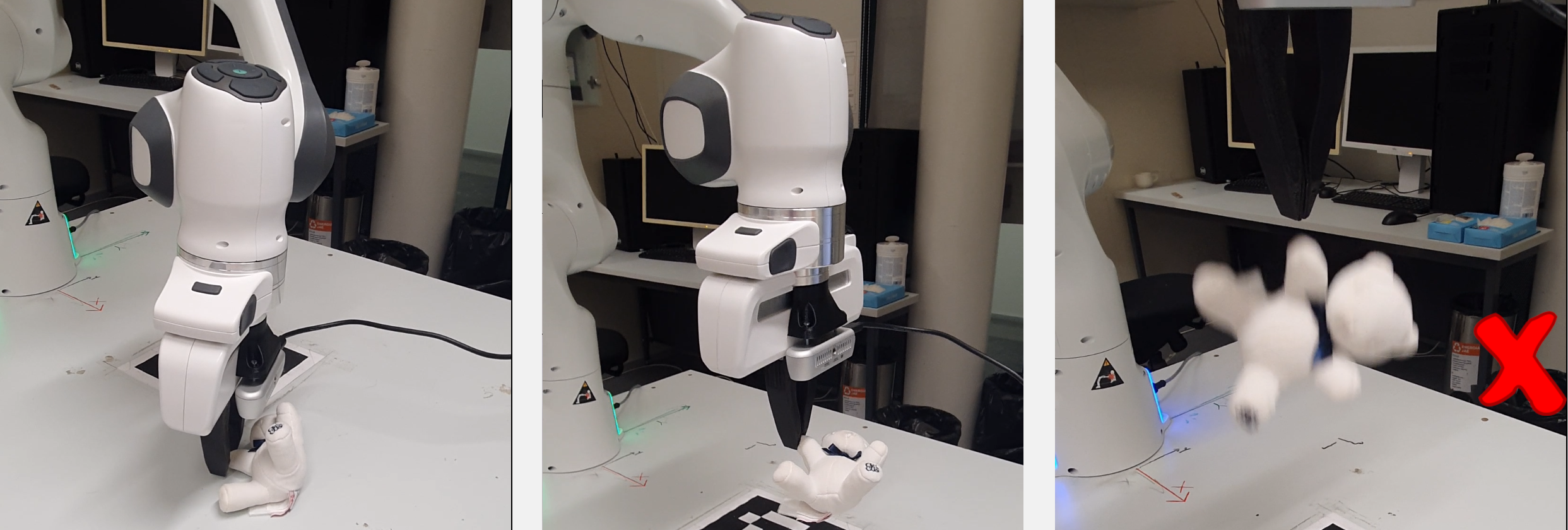}  \\
	\vspace{-0.5 em}
    \subfloat[]{
    	\includegraphics[width=.9\linewidth]{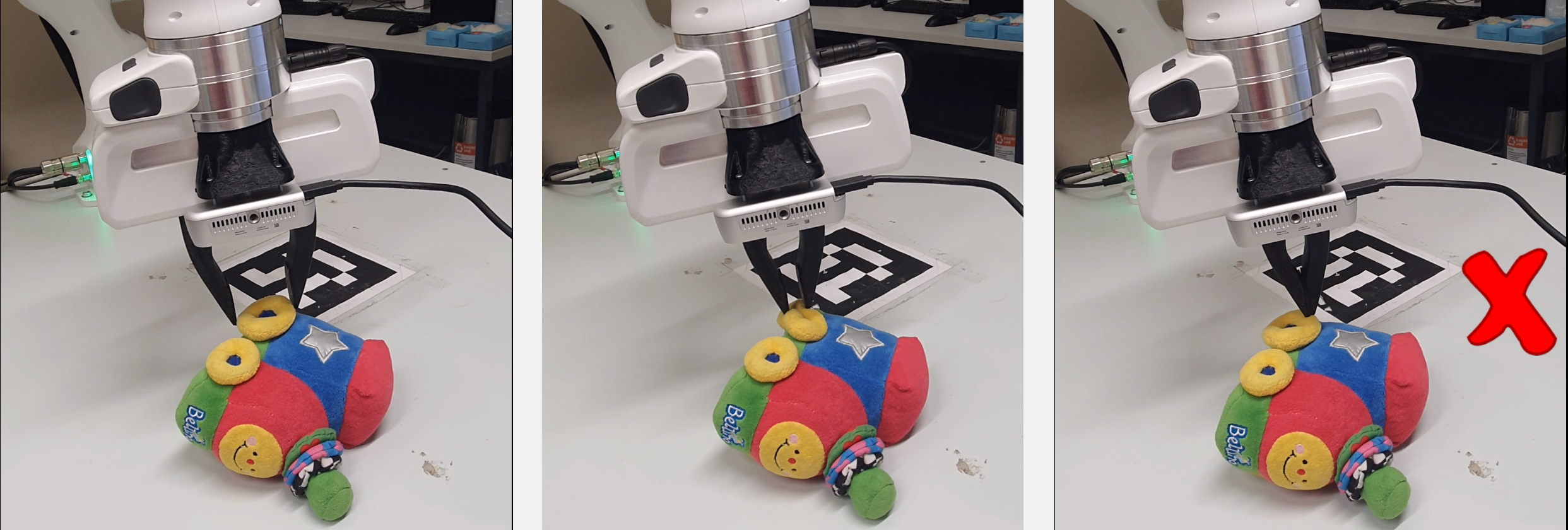} } \\
\caption{Some successful (a) and failed (b) grasps on the real robot.}
\label{fig:failandsuccess}
\vspace{-0.5cm}
\end{figure}

For instance, if we assumed the deformable objects, such as objects 8, 9, 11, and 14, were rigid, many generated grasps were on specific parts of them, such as the arms or legs of the toy or wheel of the car. These grasps usually picked the object successfully but then, due to the deformation, dropped it when the robot started to accelerate. The same experiment on objects 6, 7 showed that grasps generated on the top or bottom of the object usually failed due to the elasticity of the object. If we instead assumed rigid objects, such as 1, 3, 4, 5, were soft, the network generated pinch grasps that failed due to collisions with the object. Some failed grasps are shown in \figref{fig:failandsuccess}a. 


One interesting finding was that given the correct stiffness information, the method was able to generate different grasping strategies depending on the stiffness of the object, as shown in \figref{fig:graspstrategies}. Specifically, in the case of the soft sponge \figref{fig:graspstrategies}a, the proposed method learned that the grasp quality is high across the whole objects thanks to their deformation, which in turn, enables pinch grasps. While in the case of a hard sponge \figref{fig:graspstrategies}b, the high-quality grasp tends to be generated at the center of the object, and the grasp width is almost as big as the object in order to successfully cage the object.

\subsection{Discussion}

All experimental results show the benefit of generating stiffness-aware grasps. By comparing the proposed approach to the case where the stiffness information is ignored, we see that the proposed approach achieves higher grasp success rates. The primary reason for the difference in performance is that the object stiffness facilitates learning where to generate grasps that minimize the slippage caused by the deformation. If the object's stiffness was ignored, the network generated the same grasps regardless of the object stiffness. Together, these result backs the claim made in \cite{lin_feel3d} that grasps do not transfer well between rigid and deformable objects. Therefore, incorporating object stiffness in robotic grasping pipelines is beneficial when dealing with a wide range of unknown objects.

Another interesting finding is that our approach can generate different grasp types such as pinch or cage grasps depending on the stiffness, even though there was no pinch grasps in the training dataset. This behavior is shown on object 6 in \figref{fig:graspstrategies}. Specifically, the sponge with a low Young's modulus admits pinching behavior where the grasp press on the object and pinch, while the hard sponge only admits caging grasps. One potential reason for such behavior is that the proposed network learned that the grasp quality is almost uniform across soft objects thanks to their deformation. Similar behaviors were also reported in \cite{sergey_handeye} where data was collected from 14 robots over the course of two months. However, it is worth pointing out that our approach learned to produce the same behavior on a completely synthetic dataset with orders of magnitude fewer data. Furthermore, our proposed approach provides more meaningful insights regarding the relationship between object deformation and grasps.

\section{Conclusions and future work}
\label{sec:conclusions}
Grasping deformable objects has not been as well studied as rigid object grasping due to complexity in the modeling and simulating the dynamic behavior of such objects. However, with the rapid development of physics-based simulators that support soft bodies, the research gap between rigid and deformable objects is shrinking. To leverage the capability of such simulators and to challenge the rigidity assumption that has dominated  robotic grasping, we presented an approach to synthesize grasps on objects with varying stiffness by a deep neural network trained on purely synthetic data. The key idea in this work is to integrate the object stiffness property into the grasp planning pipeline to study the relationship between the object deformation and the generated grasps. To train the proposed network, we generated our own training dataset using the Isaac Gym simulator. We demonstrated the performance of the generated grasps through experiments in both simulation and real-world scenarios on a wide range of objects with varying sizes, shapes, and stiffness. The results show a clear improvement in grasp success rate when taking stiffness property into account. Furthermore, the proposed approach shows the ability to generate different grasp strategies depending on object stiffness. The generalizability to objects with non-uniform stiffness remains open although the method should be able to account for all variability captured in the training data, making the simulation quality and efficiency a central bottleneck.



The idea of exploiting deformations for grasping opens many interesting avenues. Our proposed method employs FEM simulations for which the computational cost may limit their use for general-purpose systems. Investigating the possibility to devise analytical quality measures that would exploit the deformations is an important future work. 




\bibliographystyle{IEEEtran}
\bibliography{refs}

\end{document}